\documentclass[letterpaper]{article} 
\usepackage{aaai25}
\usepackage{times}  
\usepackage{helvet}  
\usepackage{courier}  
\usepackage[hyphens]{url}  
\usepackage{graphicx} 
\usepackage{natbib}  
\usepackage{caption} 
\usepackage{algorithm}
\usepackage{algorithmic}

\usepackage{newfloat}
\usepackage{listings}
\DeclareCaptionStyle{ruled}{labelfont=normalfont,labelsep=colon,strut=off} 
\floatstyle{ruled}
\newfloat{listing}{tb}{lst}{}
\floatname{listing}{Listing}
\nocopyright 

\usepackage{multirow}
\usepackage{amsmath}
\usepackage{subcaption}
\usepackage{amsfonts}
\usepackage{xcolor}

\title{SASE: A Searching Architecture for Squeeze and Excitation Operations}
\author{
    Hanming Wang\textsuperscript{\rm 1},
    Yunlong Li,
    Zijun Wu,
    Huifen Wang,
    Yuan Zhang
}
\affiliations{
    China Telecom Research Institute (CTRI), China \\
    wanghm4@chinatelecom.cn\textsuperscript{\rm 1}

}

\usepackage{bibentry}

\begin{document}

\maketitle

\begin{abstract}
In the past few years, channel-wise and spatial-wise attention blocks have been widely adopted as supplementary modules in deep neural networks, enhancing network representational abilities while introducing low complexity.
Most attention modules follow a squeeze-and-excitation paradigm.
However, to design such attention modules, requires a substantial amount of experiments and computational resources.
Neural Architecture Search (NAS), meanwhile, is able to automate the design of neural networks and spares the numerous experiments required for an optimal architecture.
This motivates us to design a search architecture that can automatically find near-optimal attention modules through NAS.
We propose \textbf{SASE}, a \textbf{S}earching \textbf{A}rchitecture for \textbf{S}queeze and \textbf{E}xcitation operations, to form a plug-and-play attention block by searching within certain search space.
The search space is separated into 4 different sets, each corresponds to the squeeze or excitation operation along the channel or spatial dimension.
Additionally, the search sets include not only existing attention blocks but also other operations that have not been utilized in attention mechanisms before.
To the best of our knowledge, SASE is the first attempt to subdivide the attention search space and search for architectures beyond currently known attention modules.
The searched attention module is tested with extensive experiments across a range of visual tasks.
Experimental results indicate that visual backbone networks (ResNet-50/101) using the SASE attention module achieved the best performance compared to those using the current state-of-the-art attention modules.
Codes are included in the supplementary material, and they will be made public later.
\end{abstract}

%

\section{Introduction}

Attention mechanisms are a crucial technique in computer vision, mimicking how the human visual system focuses when observing a scene.
They help models process information more effectively, thereby improve performance.
Various attention modules have been designed and applied to a wide range of tasks.
Self-attention \cite{vaswani2017attention,cao2019gcnet,liu2021swin,han2023flatten} allows models to capture global information by broadening their receptive fields and has been widely adopted in convolutional neural networks (CNNs) and transformers due to its extraordinary representational capacity.
Soft and hard attention \cite{xu2015show} can help visualize where model actually "focus on" when performing image captioning.
Channel-wise and spatial-wise attention \cite{hu2018squeeze,ruan2021gaussian,li2024ean,jiang2024mca} adopt a squeeze-and-excitation paradigm that extracts features through a squeeze operation and produces an attention map through an excitation operation.
Enhancing model performance with minimal additional parameters makes them common auxiliary modules for CNNs.

The above attention modules are typically developed based on expert experience and require multiple experimental attempts, which can be labor-intensive.
Furthermore, given that most squeeze-and-excitation-style attention blocks follow similar design patterns, we would like to explore the possibility of designing such modules in a more automatic fashion, and NAS \cite{lu2023neural} is a powerful tool to accomplish this objective.
Several works also focused on developing attention blocks using NAS.
Ma et al. \cite{ma2020auto} created a search space comprising existing attention blocks such as the SE block \cite{hu2018squeeze} and CBAM \cite{woo2018cbam}.
They formed a fused attention module by applying differentiable architecture search (DARTS) \cite{liu2018darts} to combine the attention sub-blocks.
Nakai et al. \cite{nakai2020att} adopted a similar approach by expanding a DARTS-like search space with well-known attention modules.
They searched for an architecture that incorporated these attention blocks along with different convolution and pooling operations.
The above two methods demonstrated that a well-searched combination of different attention modules can indeed outperform its individual components.

However, the search space of existing algorithms are limited to off-the-shelf attention blocks.
Even the best searched architectures are merely combinations of existing attention sub-modules.
This inflexible search space restricts the capability of the entire search architecture, preventing the exploration of diverse potential solutions and thus limiting the performance of the searched attention module.

To address this shortcoming, this paper proposes SASE: a \textbf{S}earching \textbf{A}rchitecture for \textbf{S}queeze and \textbf{E}xcitation operations. 
In SASE, squeeze and excitation operations in the channel and spatial dimensions are searched separately in four different sets.
Operations within each set can be drawn from a diverse range of sources, including both existing attention blocks and novel techniques.
These encompass squeeze or excitation operations from existing attention mechanisms (such as global average pooling from SE-Net \cite{hu2018squeeze}, second-order pooling from GSoP-Net \cite{gao2019global}, and one-dimensional convolution from ECA-Net \cite{wang2020eca}), as well as operations not previously utilized in attention before (like \(L_p\)-pooling \cite{sermanet2012convolutional} and rank-based pooling \cite{shi2016rank,dong2017cunet}).
Searching within our proposed search space yields possible squeeze-and-excitation combinations distinct from existing attention modules.
Additionally, SASE is easily scalable by incorporating new squeeze or excitation operations into each search set.
We summarize the main contributions of our work as follows:

\begin{itemize}
    \item To enable a more fine-grained searching, we designed four distinct operation sets instead of one search set: channel-wise squeezing, channel-wise excitation, spatial-wise squeezing, and spatial-wise excitation.
    Additionally, we modified certain operations within each set to ensure consistent output dimensions.
    \item We constructed a Directed Acyclic Graph (DAG) tailored to the commonly used squeeze-and-excitation paradigm as the search starting point, with each edge corresponding to one of the four operation sets.
    And, we applied second-order DARTS to the customized DAG for an efficient search.
    \item To validate the effectiveness of SASE, the searched attention module was integrated into ResNet-50 and ResNet-101 backbones and evaluated on various vision benchmarks: COCO benchmarks for object detection and instance segmentation, as well as the ImageNet-1K classification benchmark.
    Results indicate that visual backbone using the SASE attention module achieved the best overall performance compared to those using the current state-of-the-art attention modules.
\end{itemize}

\section{Related Work}

\textbf{Attention Mechanisms}\indent Since Hu et al. \cite{hu2018squeeze} proposed SE-Net, numerous attention blocks were developed following this squeeze-and-excitation paradigm.
SE-Net utilizes a global average pooling (GAP) as the squeeze module to extract global information from each channel and applies fully-connected (FC) layers as the excitation module to analyze the inter-channel relationships and generates an attention map to re-weight the original feature.

CBAM \cite{woo2018cbam} adopts global max pooling (GMP) along with GAP to obtain more comprehensive global context.
Following similar approaches, SRM \cite{lee2019srm} incorporated global standard deviation into the extracted global features, and MCA \cite{jiang2024mca} makes use of the third-order moment (skew) to ensure a richer probability representation.
Treating GAP as a generalized \(L1\)-normalization, GCT-Net \cite{yang2020gated} applies \(L2\)-normalization as the squeeze module because GAP might fail in some extreme cases (for example, after an Instance Normalization layer).
EAN \cite{li2024ean} extends GCT-Net by separating the input tensor into several groups and individually extracting \(L2\)-norm within each group.
EPSA-Net \cite{zhang2022epsanet} also adopts the grouping mechanism by extracting global information through grouped convolutions.
GSoP-Net \cite{gao2019global} replaces the GAP with a global second-order pooling to model the correlations between channels.

Other than improving the squeeze module, some works also focused on enhancing the excitation module.
ECA-Net \cite{wang2020eca} substitutes FC layers with a 1-dimensional convolution as a light-weight alternative.
Also aiming at reducing excitation complexity, GCT-Net \cite{yang2020gated} and EAN \cite{li2024ean} assign a set of learnable scaling and shifting factor for each channel.
BAT-Net \cite{fang2019bilinear} proposes an attention-in-attention mechanism and applies an SE block as the excitation module.

\noindent\textbf{Neural Architecture Search}\indent Traditionally, designing neural architectures is a labor-intensive work.
It often requires extensive expertise and numerous trials to arrive at an optimal configuration.
NAS, however, automates this process by leveraging evolutionary algorithms, reinforcement learning, and other advanced techniques to efficiently explore the vast search space.
Over the years, various search strategies were developed, demonstrating notable enhancements in both performance and search efficiency.

Earlier works mainly focused on reinforcement learning \cite{zhong2018practical,zoph2018learning} and evolutionary algorithms \cite{real2019aging,real2019regularized}.
While achieving comparable or even superior performance to hand-crafted neural networks, the search time and computational resource requirements remain impractically high.
This issue was addressed by Differentiable Neural Architecture Search (DNAS) where the resulting architecture was obtained through gradient decent from a super-network containing all the possible candidate networks.
One of the most famous algorithms in this category is DARTS \cite{liu2018darts}.
It transforms architecture selection into a continuous problem by assigning learnable coefficients to candidate operations and selecting operations based on the values of these coefficients.
The learnable coefficients are trained together with the super-network in a two-step fashion, where in each iteration, weights in the super-network are updated first and fixed before updating the coefficients.
DARTS attained state-of-the-art (SOTA) performance on cifar-10 with only a few GPU days of searching.
A few works improved DARTS by tackling its over-representation of skip connections \cite{chu2020darts,ye2022b,zhang2023differentiable}.



\begin{figure*}[t]
\centering
\includegraphics[width=0.95\textwidth]{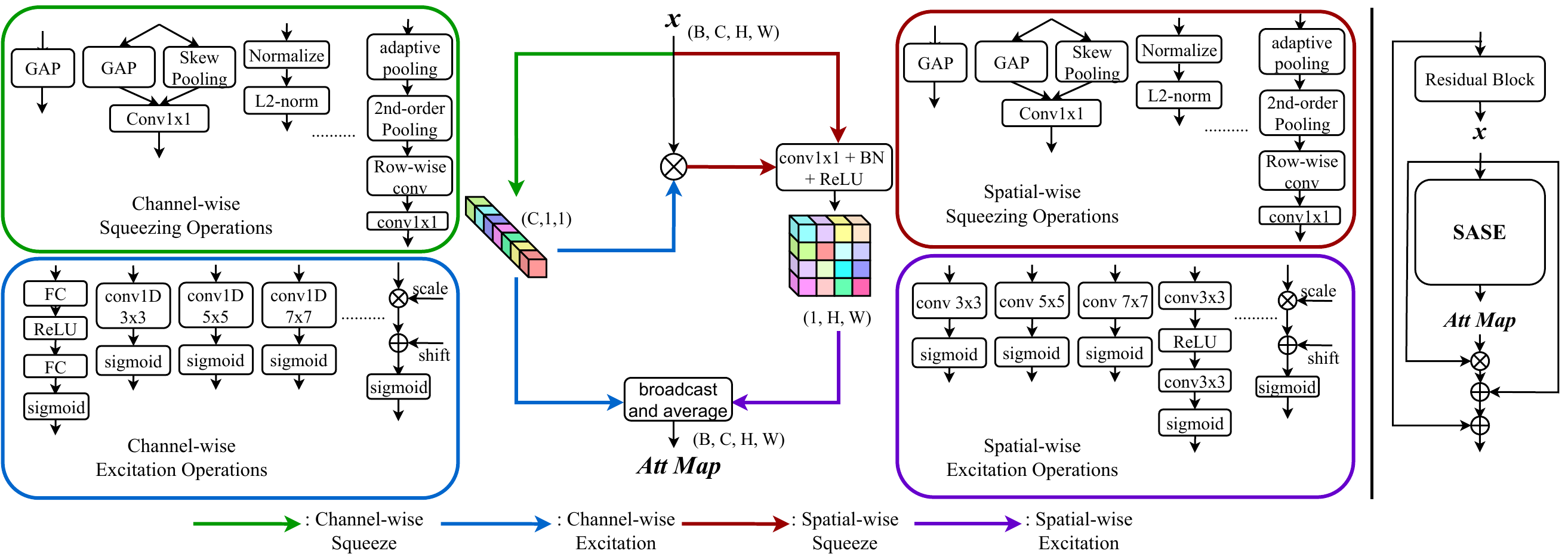} 
\caption{Overall searching architecture and search space of SASE}
\label{generalArch}
\end{figure*}

\section{The Proposed Approach}

\subsection{Search Architecture}
Since most channel-wise and spatial-wise attention blocks are designed following the same squeeze-and-excitation pattern, developing an attention architecture can be formulated as a NAS problem.
We adopt the DARTS paradigm by constructing a super-network and develop the optimized architecture through gradient decent.
The search space of designing a one-branch attention module like SE-Net \cite{hu2018squeeze} or ECA-Net \cite{wang2020eca} with NAS is too restrictive in size, which makes search trivial.
Therefore, we expand the search space by introducing a more complex super-network in the form of a DAG with each edge corresponding to one of the four operation sets: Channel-wise squeezing, channel-wise excitation, spatial-wise squeezing, and spatial-wise excitation.

As shown in Figure \ref{generalArch}, the searching architecture takes a feature tensor of size
\((B, C, H, W)\)  as input and outputs an attention map of the same size.
\(B, C, H, W\) stands for batch size, channel counts, height, and width of the input tensor respectively.
The attention map is generated by two branches: channel-wise attention and spatial-wise attention.
The former outputs a tensor of size \((B, C, 1, 1)\), while the latter produces a tensor of dimension \((B, 1, H, W)\). 
The two attention maps are first broadcast to match the input feature dimensions and then averaged to produce the final output:
\begin{equation*}
SASE\left(x\right)=\frac{attn_{ch,1}\left(x\right)+attn_{sp}\left(x, attn_{ch,2}\left(x\right)\right)}{2},
\end{equation*}
where \(x \in \mathbb{R}^{B \times C \times H \times W}\) stands for the input tensor, \(attn_{ch,1}\left(\cdot\right)\) denotes the channel-wise attention branch, and \(attn_{sp}\left(\cdot, \cdot\right)\) represents the spatial-wise attention branch.
\(attn_{ch,2}\left(\cdot\right)\) denotes an intermediate attention operation that will be introduced in the next paragraph.

The \textbf{channel-wise attention branch} contains a squeezing edge and two excitation edges.
The squeezing edge extracts channel-wise information through all the candidate operations and forms a compact implicit representation with size \((B, C, 1, 1)\).
The same channel-wise information is fed to two excitation edges to generate two attention maps.
The first attention mask directly serve as the output of channel attention branch.
The second mask is multiplied by the input tensor and then proceeds to the spatial branch for further processing:
\begin{equation*}
attn_{ch,i}\left(x\right)=excitation_{ch,i}\left(squeeze_{ch}\left(x\right)\right)),
\end{equation*}
where \(i \in \{1,2\}\) stands for the edge index, and \(excitation_{ch,i}\left(\cdot\right)\) represents the \(i^{th}\) channel-wise excitation edge.
\(squeeze_{ch}\left(\cdot\right)\) denotes the common channel-wise squeezing edge shared across two excitation edges.

The \textbf{spatial-wise attention branch} contains two squeezing edges and one excitation edge.
The first squeezing edge extracts spatial-wise information from the input tensor while the second one operates on the channel-wise-attended input.
The output to both squeezing edges are of size \((B, 1, H, W)\).
Then, the extracted spatial representation are concatenated along the channel dimension and combined by a convolution with kernel size one, a batch normalization (BN), and a ReLU non-linearity.
Lastly, the excitation edge takes the combined spatial information, with size \((B, 1, H, W)\), as input and generates the spatial-wise attention map:
\begin{multline*}
attn_{sp}\left(x, attn_{ch,2}\left(x\right)\right)= \\
excitation_{sp}\left(conv_{1\times1}\left(interm\left(x,attn_{ch,2}(x)\right)\right)\right),
\end{multline*}
where \(excitation_{sp}\left(\cdot\right)\) denotes the spatial-wise excitation edge, and \(conv_{1\times1}\left(\cdot\right)\) represents the convolution block containing convolution with kernel size one, BN, and ReLU.
\(interm\left(\cdot\right)\) stands for the channel-wise concatenation of two squeezing outputs:
\begin{multline*}
interm\left(x,attn_{ch,2}\left(x\right)\right)= \\
concat\left(squeeze_{sp,1}\left(x\right),squeeze_{sp,2}\left(attn_{ch,2}\left(x\right) \otimes x\right)\right),
\end{multline*}
where \(squeeze_{sp,1}\left(\cdot\right)\) denotes the spatial-wise squeezing edge operating on the input tensor, while \(squeeze_{sp,2}\left(\cdot\right)\) denotes the one operating on the channel-wise re-calibrated feature: \(attn_{ch,2}\left(x\right) \otimes x\).

\subsection{Search Space}
Each aforementioned edge in the DAG corresponds to an operation set that contains 7 operations.
In this section, we introduce each set and each operation in details.

\subsubsection{Channel-wise Squeezing Set}

The channel-wise squeezing set is comprised of operations that extract global spatial information and generate one value per channel that encapsulates the channel dynamics. This set contains the following operations:

\textbf{\textit{Global Average Pooling (GAP)}} was first introduced by SE-Net \cite{hu2018squeeze} to extract global context.
We adopted GAP into the operation set because of its effectiveness in capturing global spatial information.

\textbf{\textit{Global Second-order Pooling (GSoP-Net)}} \cite{gao2019global} proposed to adopt second-order pooling (also known as covariance pooling) to capture correlation between channels.
Since calculating correlations between each channel pair requires an inner product of two vectors with size \(\mathrm{H} \times \mathrm{W}\), which is time consuming, we apply a channel-wise adaptive average pooling to shrink the channel count by a factor of 4 before the covariance pooling.
In order to perform a DARTS-style searching, the output dimensions of operations within each set needs to be the same.
To achieve this, we first apply a row-wise convolution after the pooling to compress the two-dimensional channel-wise correlation of size \(\frac{C}{4} \times \frac{C}{4}\) into one dimension.
Next, a convolution with kernel size 1 is applied to adjust the channel numbers from \(\frac{C}{4}\) back to \(C\).

\textbf{\textit{Standardization and L\textsubscript{2}-normalization}}: \(L_2\)-normalization was introduced by GCT-Net \cite{yang2020gated} and EAN \cite{li2024ean} as a different method of information extraction.
Following EAN \cite{li2024ean}, we implemented an instance normalization, which applies a standardization and an affine transform along the spatial dimension, before \(L_2\)-normalization to smoothen spatial statistics.

\textbf{\textit{L\textsubscript{4}-pooling}}: \(L_p\)-pooling was first proposed by Sermanet et al. \cite{sermanet2012convolutional} as an alternate to average pooling and max pooling.
\(L_p\) pooling for a given channel takes a form of :
\begin{equation*}
\left(\sum_{i=0}^H \sum_{j=0}^W x_{(i,j)}^p\right)^{1/p},
\end{equation*}
where \(x_{(i,j)}\) denotes each spatial location of the input tensor, and \(p\) is a hyper-parameter to be chosen.
As \(p\) approaches infinity, \(L_p\)-pooling approximates max pooling, because the influence of smaller values becomes negligible.
When \(p\) equals to one, \(L_p\)-pooling corresponds to average pooling, as it sums the values and divides by the number of elements.
We observed that if global max pooling (GMP) is separately included in the squeezing set, it will almost never be chosen because it limits the gradient update to the largest number in each channel.
Therefore, we approximate GMP by applying a \(L_4\)-pooling.
During our experiments, any \(p\) larger than 4 caused gradients explosion, and thus choosing \(p=4\) is sufficient to amplify the impact of the maximum value on global spatial context.
Unlike \(L_2\)-normalization, standardization is not applied since it would suppress the largest value and diminish the effect of \(L_4\)-pooling.

\textbf{\textit{Global Average Pooling and Global Max Pooling}}: CBAM \cite{woo2018cbam} added a parallel GMP to complement GAP for a richer global context extraction.
Adopting similar design, we separately applied GAP and GMP to the original feature tensor with each squeezing operation extracting a tensor of size \((B, C, 1, 1)\).
To align the output channel count to \(C\), we combine the extracted global spatial information using a convolution block.

\textbf{\textit{Global Average Pooling and Standard Deviation Pooling}}: Similar to CBAM, SRM \cite{lee2019srm} added a parallel standard deviation pooling to enrich the global context.
We implement this module in the same way as the "GAP and GMP" squeezing block.

\textbf{\textit{Global Average Pooling and Skew Pooling}}: Similar to the two aforementioned squeezing operations, MCA \cite{jiang2024mca} adopts skew pooling to accompany GAP and makes use of the third-order moment.
This module is also implemented in the same way as the two modules introduced above.

\subsubsection{Spatial-wise Squeezing Set}
The spatial-wise squeezing set contains the same type of operations as in the channel-wise squeezing set, except for the dimension that squeezing operates on.
Each operation gathers channel information on a global scale and generates one value per spatial location that reflects channel statistics at that point.
All the pooling, normalization, standardization and convolution blocks in spatial-wise squeezing modules operates at the channel level, unlike its channel-wise squeezing counterpart, which operates on the height and width dimensions.
Since only the dimension is switched, details on implementation of each operation will not be further elaborated.

\subsubsection{Channel-wise Excitation Set}
Operations in excitation sets are not as diverse as the ones in squeezing sets.
They are mainly composed of convolutions with different hyper-parameters, fully-connected (FC) layers, and affine transforms.
These operations aggregate the extracted global information and generate an attention map that scales the original input.
In order to bound the attention map between 0 and 1, a \(sigmoid\) non-linearity is attached to each excitation operation.
Details of each channel-wise excitation operation are introduced below:

\textbf{\textit{Fully-connected layers with reduction}}: Using FC layers as the excitation module was first introduced in SE-Net \cite{hu2018squeeze}.
Following SE-Net, we used two layers of FC layer with the first one reducing the feature channel count by a factor of 16 and the second one restoring feature channels back to its original number.

\textbf{\textit{1D-convolutions with kernel size 3, 5, and 7}}: ECA-Net \cite{wang2020eca} adopts a 1D-convolution (1D-conv) in replacement of the FC layers as a simplified substitute.
We included 1D-conv with kernel size 3, 5, and 7 for better flexibility of the searching algorithm to choose the most appropriate kernel size.

\textbf{\textit{Stacked 1D-convolutions with kernel size 3}}:
Since the receptive field of a convolution with large kernel size can be covered by multiple convolutions with small kernel sizes, we adopt 2 and 3 layers of 1D-conv of kernel size 3 stacking together and connected with ReLU non-linearities.

\textbf{\textit{Affine transform}}: 
Affine transform was adopted in GCT-Net \cite{yang2020gated} and EAN \cite{li2024ean} as another light-weight excitation module.
Two learnable weights are assigned to each channel to scale and shift the channel context.

\subsubsection{Spatial-wise Excitation Set}
As opposed to the channel-wise excitation edges, the input to the spatial-wise excitation edge has no channel dimension.
Hence, most excitation operations in this set are 2D single-kernel convolutions.
Detailed introduction to each spatial-wise excitation operation is shown below:

\textit{\textbf{2D-convolution with kernel size 3, 5, and 7}}:
CBAM \cite{woo2018cbam} adopts a 2D-convolution (2D-conv) of kernel size 7 as the spatial-wise excitation module.
We include 2D-convs with 3 different kernel sizes, which allows the searching algorithm to adaptively find the most suitable one.

\textit{\textbf{Stacked 2D-convolutions with kernel size 3}}:
To cover a 5x5 receptive field, we adopt 2 layers of 2D-convs with kernel size 3 stacking together and connected with ReLU non-linearities.

\textit{\textbf{Spatially separable convolution}}:
Spatially separable convolution is a technique that breaks a \(k \times k\) kernel into a \(k \times 1\) and a \(1 \times k\) kernel to reduce the number of multiplications while keeping the same receptive field.
We incorporate spatially separable convolution with kernel size 3 and 5 as a lightweight alternative of normal 2D-convs.

\textit{\textbf{Affine transform}}: 
Similar to channel-wise affine transform, spatial affine transform assigns 2 learnable weights to each spatial location for scaling and shifting.

\begin{figure}[t]
\centering
\includegraphics[width=0.9\columnwidth]{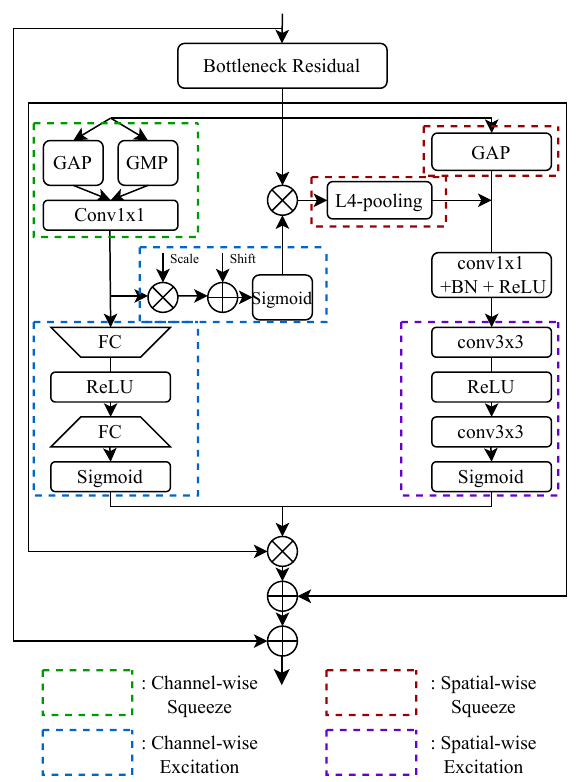} 
\caption{The searched attention module and how it was implemented in ResNet}
\label{resultingArch}
\end{figure}


\subsection{Search Process}
Having defined the super-network and search space in detail, we can now formulate a differentiable search on the architecture.
Given the super-network defined as a DAG, a learnable weight is assigned to each operation on each edge.
Operations on a given edge are combined in the following way:
\begin{equation*}
\mathrm{edge}_{i}(x)=\sum_{o \in \mathcal{O}} \frac{\exp \left(\alpha_{i,o}\right)}{\sum_{o^{\prime} \in \mathcal{O}} \exp \left(\alpha_{i,o^{\prime}}\right)} o(x),
\end{equation*}
where \(\mathrm{edge}_{i}(\cdot)\) denotes the \(i^{th}\) edge in the DAG, and \(x\) represents the input to the given edge.
\(\mathcal{O}\) denotes one of the four operation sets introduced in the Search Space sub-section, and \(o \in \mathcal{O}\) stands for a candidate operation in a given set.
\(\alpha_{i,o}\) is the weight assigned to operation \(o\) on \(\mathrm{edge}_{i}\).
Softmax is calculated over all the operations within each set so that the coefficients add up to one.
When the search converges, the optimal operation at \(\mathrm{edge}_{i}\) is chosen according to its corresponding weight:
\begin{equation}
\mathrm{o}^{*}_{i}=\mathrm{argmax}_{\mathrm{o} \in \mathcal{O}} \: \alpha_{i,o}.
\end{equation}

The coefficient set \(\alpha\) and the weight of the super-network \(\omega\) are jointly learned  during training.
Following DARTS \cite{liu2018darts}, the joint training is formulated as a bi-level optimization problem:
\begin{equation} \label{biOpti}
\begin{gathered}
\min _\alpha \mathcal{L}_{val}\left(\omega^*(\alpha), \alpha\right) \\
\text { s.t. } \omega^*(\alpha)=\underset{\omega}{\operatorname{argmin}} \mathcal{L}_{train}\left(\omega, \alpha\right),
\end{gathered}
\end{equation}
Where \(\mathcal{L}_{train}\) and \(\mathcal{L}_{val}\) each stands for the task loss on training and validation set, respectively.
However, calculating the exact gradients from Eq. (\ref{biOpti}) for the searching architecture requires training the super-network till convergence at every optimization step, which is not realistic.
Liu et al. \cite{liu2018darts} proposed a workaround to this problem by training the super-network for one step and fixing the network weight while updating parameters from the searching architecture.
\(\omega^*(\alpha)\) in Eq. (\ref{biOpti}) can therefore be replaced by:
\begin{equation} \label{aprox}
\omega^*(\alpha) \approx \omega - \eta \nabla_\omega \mathcal{L}_{train}(\omega, \alpha),
\end{equation}
where \(\eta\) is the learning rate.
The final updating equation for alpha can be derived as follows:

Treating \(\mathcal{L}_{val}\left(\tilde{\omega}, \alpha\right)\) as a multi-variable function, we can apply chain rule to the derivative:
\begin{multline} \label{chainRule}
\nabla_\alpha \mathcal{L}_{val}\left(\tilde{\omega}, \alpha\right) \\
= \nabla_\alpha \tilde{\omega} \cdot \nabla_{\tilde{\omega}} \mathcal{L}_{val}\left(\tilde{\omega}, \alpha\right) + \nabla_\alpha \alpha \cdot \nabla_\alpha \mathcal{L}_{val}\left(\tilde{\omega}, \alpha\right) \\
= \nabla_\alpha \left(\omega - \eta \nabla_\omega \mathcal{L}_{train}\left(\omega, \alpha\right)\right) \cdot \nabla_{\tilde{\omega}} \mathcal{L}_{val}\left(\tilde{\omega}, \alpha\right) \\ + \nabla_\alpha \alpha \cdot \nabla_\alpha \mathcal{L}_{val}\left(\tilde{\omega}, \alpha\right) \\
=-\eta \nabla_{\alpha, \omega}^2 \mathcal{L}_{train}(\omega, \alpha) \cdot \nabla_{\tilde{\omega}} \mathcal{L}_{val}\left(\tilde{\omega}, \alpha\right) + \nabla_\alpha \mathcal{L}_{val}\left(\tilde{\omega}, \alpha\right),
\end{multline}
Where \(\tilde{\omega}\) denotes \(\omega\) after one step of update in Eq. (\ref{aprox}).

To avoid the expensive calculation of the Hessian matrix in Eq. (\ref{chainRule}), the derivative can be approximated using finite difference:
\begin{multline}
\nabla_{\alpha, \omega}^2 \mathcal{L}_{train}(\omega, \alpha) \approx \\
\frac{\nabla_\alpha \mathcal{L}_{train}\left(\omega+h, \alpha\right)-\nabla_\alpha \mathcal{L}_{train}\left(\omega-h, \alpha\right)}{2h},
\end{multline}
as long as \(h=\epsilon\nabla_{\tilde{\omega}} \mathcal{L}_{\text {val }}\left(\tilde{\omega}, \alpha\right)\) and the scaling factor \(\epsilon\) are small enough numbers.

\begin{table*}[t]
\centering
\begin{tabular}{c|c|c|cccccc}
\hline Detector & Methods & Params & mAP & $AP_{0.5}$ & $AP_{0.75}$ & $AP_S$ & $AP_M$ & $AP_L$ \\
\hline \multirow{18}{*}{ Faster RCNN } & ResNet-50 & 41.5 M & 37.4 & 58.1 & 40.4 & 21.2 & 41.0 & 48.1 \\ 
& +SRM (ICCV,2019) & 41.6 M & 38.8 & 60.2 & 42.1 & 22.6 & 43.1 & 49.5 \\
& +GSoP (CVPR,2019) & 44.1 M & 39.2 & 60.5 & 42.6 & 23.2 & 43.1 & 49.8 \\
& +ECA (CVPR,2020) & 41.5 M & 38.0 & 60.6 & 40.9 & 22.4 & 42.1 & 48.0 \\
& +AutoLA (NeurIPS,2020) & 48.6 M & 39.1 & 60.6 & 42.7 & 23.6 & 43.0 & 49.9 \\
& +GCT (CVPR,2021) & 41.5 M & 38.9 & 60.4 & 42.3 & 22.8 & 43.1 & 49.7 \\
& +MCA (AAAI,2024) & 41.6 M & 38.3 & 60.5 & 41.4 & 22.6 & 42.4 & 49.4 \\
& +EAN (AAAI,2024) & 41.5 M & 39.0 & 60.9 & 42.6 & 23.4 & 42.7 & 50.0 \\
& +\textbf{SASE (Ours)} & 44.3 M & \textbf{39.4} & 60.6 & 43.0 & 24.0 & 43.3 & 50.4 \\
\cline{2-9}
& ResNet-101 & 60.5 M & 39.4 & 60.1 & 43.1 & 22.4 & 43.7 & 51.1 \\
& +SRM (ICCV,2019) & 60.6 M & 40.6 & 61.8 & 44.5 & 23.1 & 45.2 & 52.5 \\
& +GSoP (CVPR,2019) & 64.9 M & 41.4 & 62.4 & 45.1 & 24.7 & 45.0 & 53.6 \\
& +ECA (CVPR,2020) & 60.5 M & 40.3 & 62.9 & 44.0 & 24.5 & 44.7 & 51.3 \\
& +AutoLA (NeurIPS,2020) & 73.8 M & 41.5 & 62.6 & 45.4 & 24.9 & 45.5 & 54.1 \\
& +GCT (CVPR,2021) & 60.5 M & 40.7 & 61.9 & 44.6 & 23.3 & 45.3 & 52.8 \\
& +MCA (AAAI,2024) & 60.6 M & 40.3 & 61.9 & 43.7 & 23.5 & 44.4 & 52.5 \\
& +EAN (AAAI,2024) & 60.5 M & 41.1 & 62.2 & 44.9 & 24.5 & 45.1 & 53.3 \\
& +\textbf{SASE (Ours)} & 65.6 M & \textbf{41.8} & 62.9 & 45.8 & 25.1 & 45.7 & 54.4 \\
\hline
\multirow{18}{*}{ Mask RCNN } & ResNet-50 & 44.2 M & 38.2 & 58.8 & 41.4 & 21.9 & 40.9 & 49.5 \\
& +SRM (ICCV,2019) & 44.2 M & 39.2 & 60.9 & 42.9 & 23.4 & 43.1 & 50.5 \\
& +GSoP (CVPR,2019) & 48.6 M & 39.7 & 61.4 & 43.4 & 23.8 & 43.7 & 51.4 \\
& +ECA (CVPR,2020) & 44.2 M & 39.0 & 61.3 & 42.1 & 24.2 & 42.8 & 49.9 \\
& +AutoLA (NeurIPS,2020) & 51.3 M & 39.8 & 61.3 & 43.3 & 23.5 & 43.5 & 51.6 \\
& +GCT (CVPR,2021) & 44.2 M & 39.4 & 60.8 & 43.0 & 23.6 & 43.3 & 50.7 \\
& +MCA (AAAI,2024) & 44.2 M & 39.0 & 60.9 & 42.7 & 23.3 & 42.9 & 50.5 \\
& +EAN (AAAI,2024) & 44.2 M & 39.6 & 61.0 & 43.1 & 23.7 & 43.6 & 51.1 \\
& +\textbf{SASE (Ours)} & 46.9 M & \textbf{40.0} & 61.3 & 43.6 & 23.9 & 43.9 & 51.7 \\
\cline{2-9}
& ResNet-101 & 63.2 M & 40.0 & 60.5 & 44.0 & 22.6 & 44.0 & 52.6 \\
& +SRM (ICCV,2019) & 63.2 M & 41.1 & 62.7 & 45.0 & 24.5 & 45.3 & 53.6 \\
& +GSoP (CVPR,2019) & 67.6 M & 41.8 & 63.0 & 45.6 & 24.6 & 45.7 & 55.0 \\
& +ECA (CVPR,2020) & 63.2 M & 41.3 & 63.2 & 44.8 & 25.1 & 45.8 & 52.9 \\
& +AutoLA (NeurIPS,2020) & 76.4 M & 41.7 & 63.1 & 45.4 & 24.5 & 45.8 & 49.9 \\
& +GCT (CVPR,2021) & 63.2 M & 41.5 & 62.6 & 45.3 & 24.1 & 45.6 & 53.9 \\
& +MCA (AAAI,2024) & 63.2 M & 41.0 & 62.8 & 44.8 & 24.9 & 45.1 & 52.9 \\
& +EAN (AAAI,2024) & 63.2 M & 41.5 & 62.8 & 45.2 & 24.3 & 45.4 & 54.1 \\
& +\textbf{SASE (Ours)} & 68.2 M & \textbf{42.0} & 63.1 & 45.8 & 24.7 & 46.1 & 55.2 \\
\hline
\end{tabular}
\caption{The comparison results of object detection on COCO val2017 with other attention modules}
\label{table1}
\end{table*}

\section{Experiments}
Experiments on SASE are two-fold.
We first implemented the whole searching architecture into ResNet-20 \cite{he2016deep} and performed searching on CIFAR-10 \cite{krizhevsky2009learning}.
After the search converged, the resulting attention block was inserted into ResNet-50 and ResNet-101 backbones and tested on ImageNet-1k \cite{russakovsky2015imagenet} and COCO dataset \cite{lin2014microsoft}.
Task results were compared with other famous attention modules, especially the ones that have overlapping operations with our search space.

\subsection{Searching on CIFAR-10}
In order to maintain the search time in an acceptable range, we implemented SASE in a shallow network and searched the architecture on a relatively simple dataset.
Specifically, we inserted SASE after the last batch normalization of each residual block from ResNet-20, and the modified ResNet-20 is trained for 50 epochs on CIFAR-10 following the bi-level optimization defined in the Search Process sub-section.
Batch size was chosen to be 128.
The training set was partitioned into two halves, one of which is used to update the network weight \(\omega\), while the other is used to update the architecture parameter \(\alpha\).
For weight \(\omega\), We applied SGD optimizer with a momentum of 0.9 and a weight decay of 0.0005.
Initial learning rate was set to 0.025 and gradually decayed to 0.0001 with cosine scheduler.
For parameter \(\alpha\), we utilised Adam optimizer \cite{kingma2014adam} with an initial learning rate of 0.0003.
\(\alpha\) was initialized through sampling from a zero-mean Gaussian distribution with standard deviation 0.001 to ensure robustness. 

After training converged, the resulting attention module was formed by picking one operation per edge with the largest \(\alpha\). 
As shown in Figure \ref{resultingArch}, the final operation for each edge are as follows: \textbf{\{}\(squeeze_{ch}\): GAP and GMP, \(excitation_{ch,1}\): FC with reduction, \(excitation_{ch,2}\): Affine, \(squeeze_{sp,1}\): GAP, \(squeeze_{sp,2}\): L\textsubscript{4}-pooling, \(excitation_{sp}\): Stacked conv\textsubscript{3x3}\textbf{\}}.
In order to test the generalization ability of the resulting architecture, we implemented the obtained attention module into deeper networks and evaluated their performance on more challenging tasks.
To be specific, the widely-used ResNet-50 and ResNet-101 were chosen as the evaluation networks.
We inserted the resulting architecture after the last batch normalization of each bottleneck residual block and tested the modified network on image classification, object detection, and instance segmentation.
Results were compared with another NAS based attention method AutoLA \cite{ma2020auto}, and other attention modules: EAN \cite{li2024ean}, MCA \cite{jiang2024mca}, GCT \cite{ruan2021gaussian}, ECA \cite{wang2020eca}, GSoP \cite{gao2019global}, and SRM \cite{lee2019srm}.

\begin{figure}[t]
  \begin{subfigure}[b]{0.2\linewidth}
    \makebox[35pt]{\raisebox{18pt}{Images}}%
    \includegraphics[width=\dimexpr\linewidth\relax]{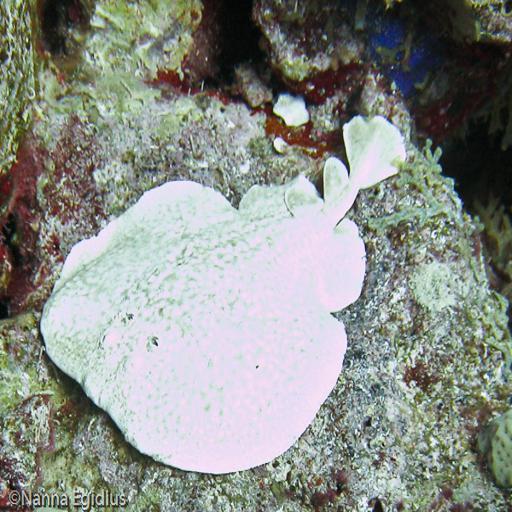}
    \makebox[35pt]{\raisebox{18pt}{Res50}}%
    \includegraphics[width=\dimexpr\linewidth\relax]{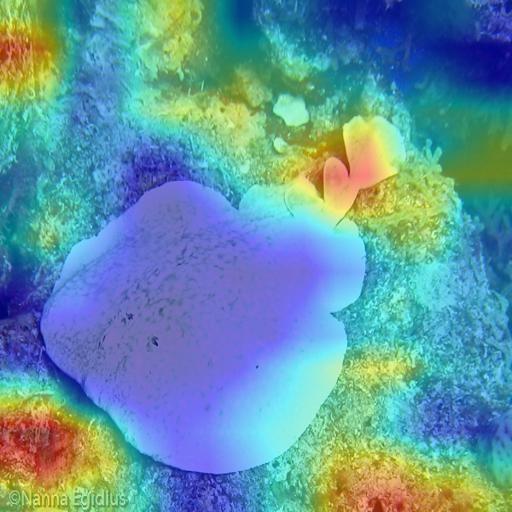}
    \makebox[35pt]{\raisebox{18pt}{CBAM}}%
    \includegraphics[width=\dimexpr\linewidth\relax]{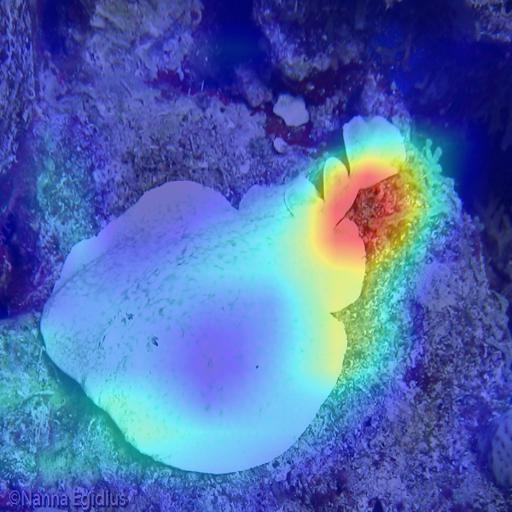}
    \makebox[35pt]{\raisebox{18pt}{ECA}}%
    \includegraphics[width=\dimexpr\linewidth\relax]{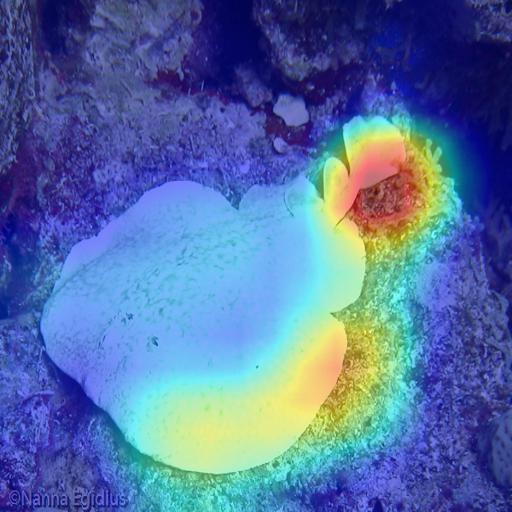}
    \makebox[35pt]{\raisebox{18pt}{SASE}}%
    \includegraphics[width=\dimexpr\linewidth\relax]{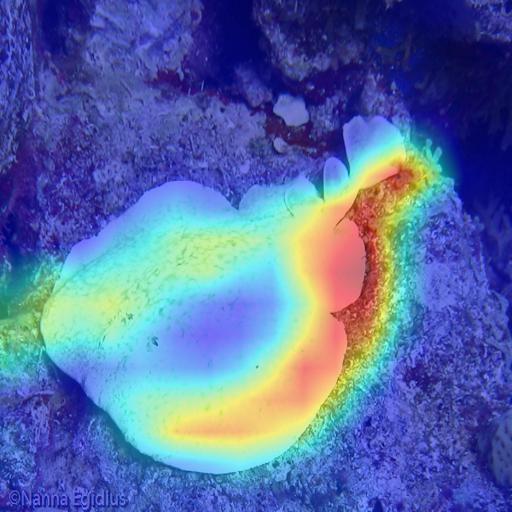}
    \makebox[35pt]{\raisebox{18pt}{}}%
    electricRay
  \end{subfigure}
  \hspace{33pt}
  \begin{subfigure}[b]{0.2\linewidth}
    \includegraphics[width=\dimexpr\linewidth\relax]{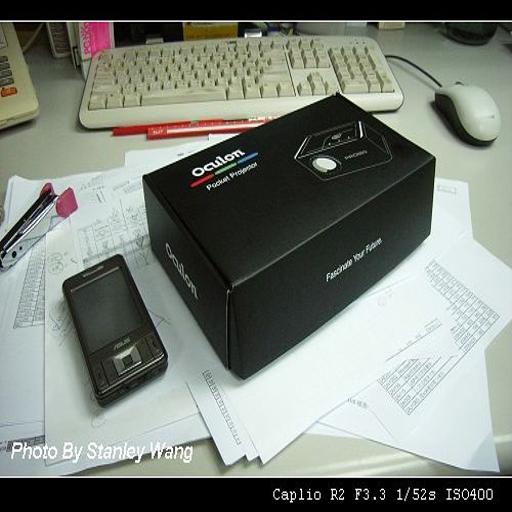}
    \includegraphics[width=\dimexpr\linewidth\relax]{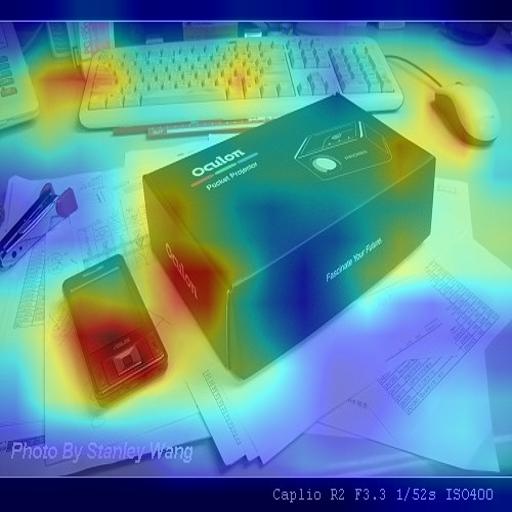}
    \includegraphics[width=\dimexpr\linewidth\relax]{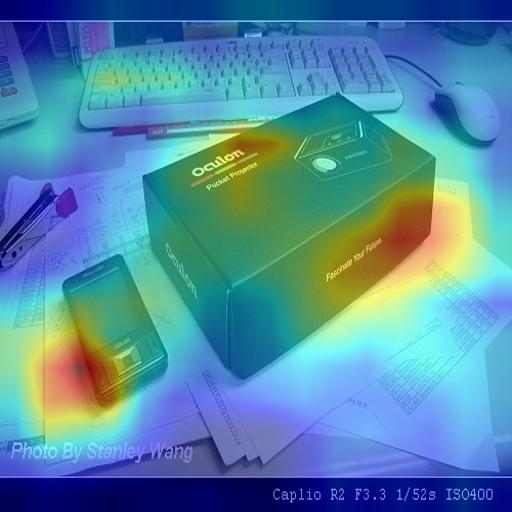}
    \includegraphics[width=\dimexpr\linewidth\relax]{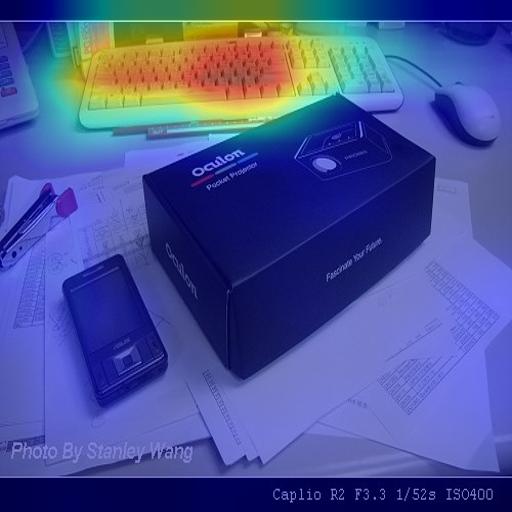}
    \includegraphics[width=\dimexpr\linewidth\relax]{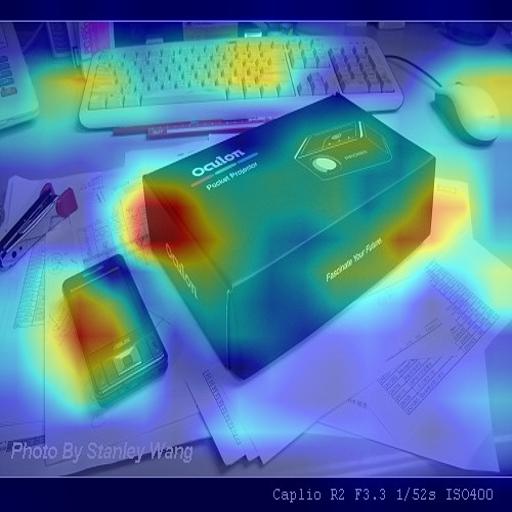}
    \makebox[6pt]{\raisebox{18pt}{}}%
    projector
  \end{subfigure}%
  \hspace{0.1pt}
  \begin{subfigure}[b]{0.2\linewidth}
    \includegraphics[width=\dimexpr\linewidth\relax]{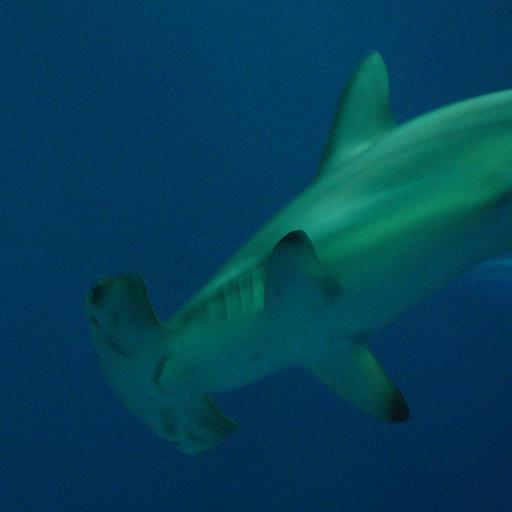}
    \includegraphics[width=\dimexpr\linewidth\relax]{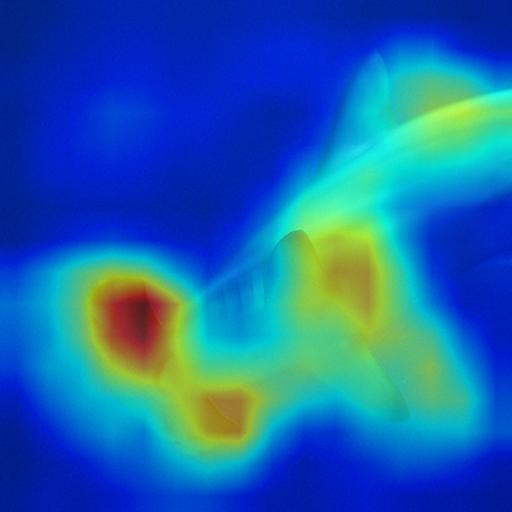}
    \includegraphics[width=\dimexpr\linewidth\relax]{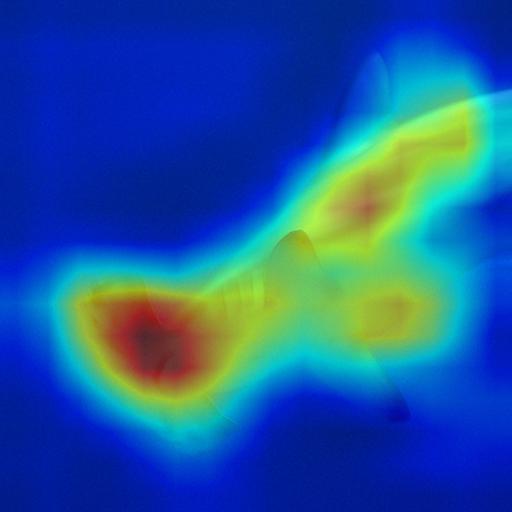}
    \includegraphics[width=\dimexpr\linewidth\relax]{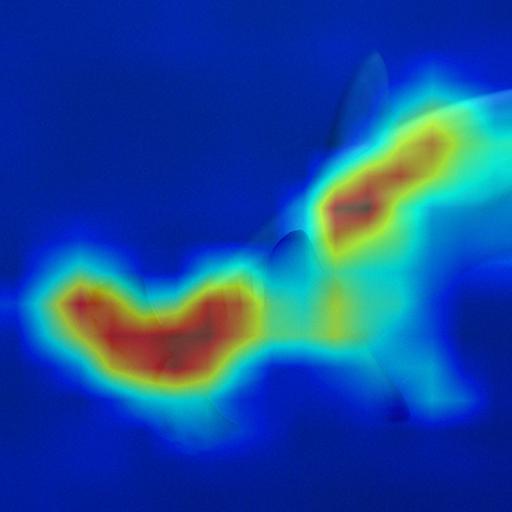}
    \includegraphics[width=\dimexpr\linewidth\relax]{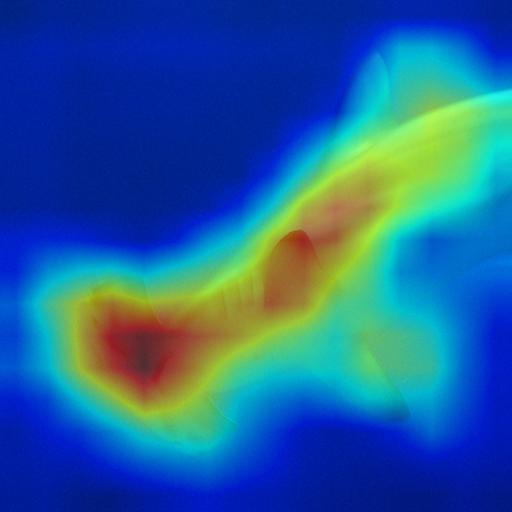}
    \makebox[0pt]{\raisebox{18pt}{}}%
    hammerhead
  \end{subfigure}%
  \hspace{0.1pt}
  \begin{subfigure}[b]{0.2\linewidth}
    \includegraphics[width=\dimexpr\linewidth\relax]{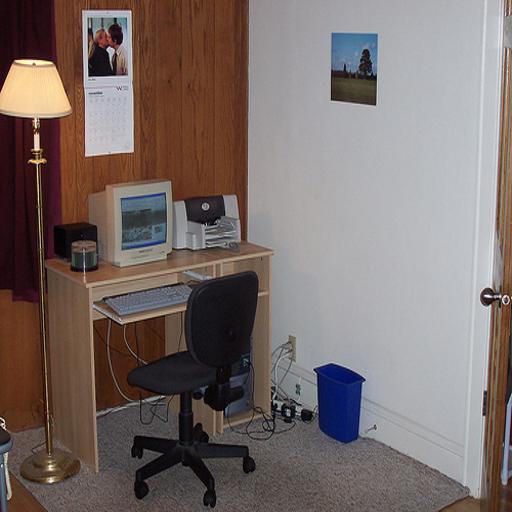}
    \includegraphics[width=\dimexpr\linewidth\relax]{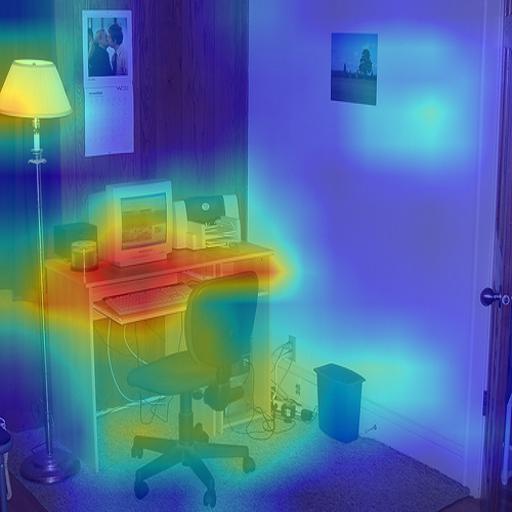}
    \includegraphics[width=\dimexpr\linewidth\relax]{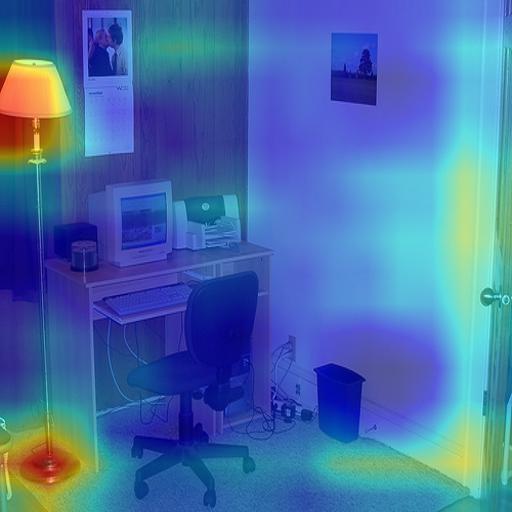}
    \includegraphics[width=\dimexpr\linewidth\relax]{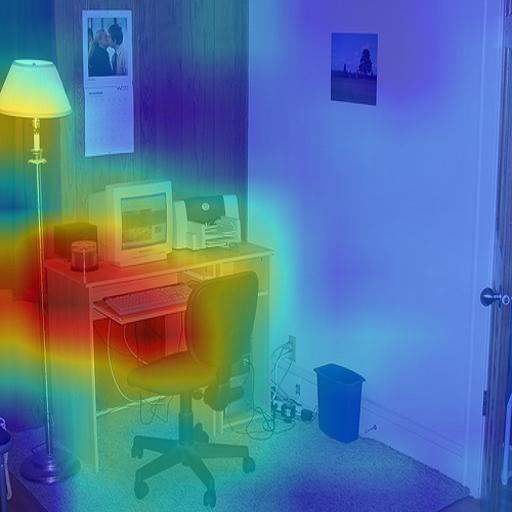}
    \includegraphics[width=\dimexpr\linewidth\relax]{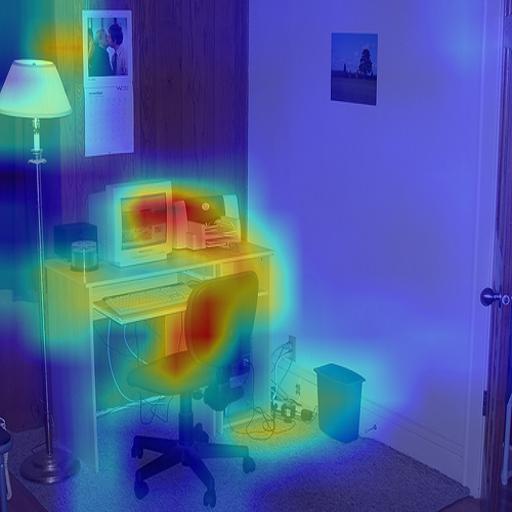}
    \makebox[8pt]{\raisebox{18pt}{}}%
    printer
  \end{subfigure}%
  \caption{Visualization of different attention modules with Grad-Cam \cite{selvaraju2017grad}}
  \label{fig:visualization}
\end{figure}

\subsection{Image Classification on ImageNet-1k}
The modified ResNet-50 and ResNet-101 were trained from scratch for 100 epochs with batch size 128.
We applied the linear scaling rule and set the initial learning rate to be 0.05, which slowly decreased to 0 following cosine scheduler.
SGD with a momentum of 0.9 and a weight decay of 0.0001 was chosen as the optimizer.
The network was trained under mobile setting where input images were randomly cropped to 224x224.
Other data augmentation settings followed the original ResNet \cite{he2016deep}.

Experimental results show that SASE outperforms all the benchmark attention modules by a significant margin on ResNet-101, and outperforms most attention modules on ResNet-50.
Although our method is on par with EAN \cite{li2024ean} and AutoLA \cite{ma2020auto} on ResNet-50, later evaluations on object detection and instance segmentation demonstrate better generalization capability of our model when applied to more challenging tasks.

\begin{table}[t]
\centering
\begin{tabular}{c|p{0.6cm}p{0.6cm}p{0.6cm}p{0.6cm}p{0.6cm}p{0.6cm}}
\hline Methods & mAP & $AP_{0.5}$ & $AP_{0.75}$ & $AP_S$ & $AP_M$ & $AP_L$ \\
\hline 
ResNet-50 & 34.7 & 55.7 & 37.2 & 18.3 & 37.4 & 47.2 \\
+SRM & 35.6 & 57.4 & 37.9 & 17.3 & 39.1 & 51.6 \\
+GSoP & 35.9 & 58.3 & 38.0 & 17.9 & 39.2 & 50.2 \\
+ECA & 35.6 & 58.1 & 37.7 & 17.6 & 39.0 & 51.8 \\
+AutoLA & 36.0 & 58.1 & 38.3 & 17.3 & 39.5 & 50.6 \\
+GCT & 35.8 & 57.5 & 38.1 & 19.8 & 39.4 & 48.2 \\
+MCA & 35.4 & 57.4 & 37.7 & 16.7 & 38.7 & 50.7 \\
+EAN & 35.9 & 58.2 & 38.2 & 17.4 & 39.0 & 50.5 \\
+\textbf{SASE} & \textbf{36.2} & 58.3 & 38.4 & 17.5 & 39.2 & 51.9 \\
\hline
ResNet-101 & 36.1 & 57.5 & 38.6 & 18.8 & 39.7 & 49.5 \\
+SRM & 37.3 & 59.8 & 39.0 & 18.3 & 41.0 & 51.9 \\
+GSoP & 37.5 & 60.0 & 39.6 & 19.2 & 41.0 & 52.5 \\
+ECA & 37.4 & 59.9 & 39.8 & 18.1 & 41.1 & 54.1 \\
+AutoLA & 37.5 & 59.9 & 39.9 & 19.7 & 41.2 & 53.2 \\ 
+GCT & 37.3 & 59.5 & 39.7 & 19.9 & 41.1 & 50.9 \\
+MCA & 36.8 & 59.5 & 38.8 & 18.0 & 40.5 & 53.3 \\
+EAN & 37.4 & 59.8 & 39.7 & 17.9 & 40.9 & 54.2 \\
+\textbf{SASE} & \textbf{37.5} & 59.8 & 39.9 & 17.8 & 41.1 & 54.8 \\
\hline
\end{tabular}
\caption{The comparison results of instance segmentation on COCO val2017 with other attention modules}
\label{table2}
\end{table}

\begin{table}[t]
\centering
\begin{tabular}{c|c|cc}
\hline Methods & Params & Top-1 acc. & Top-5 acc. \\
\hline 
ResNet-50 & 25.56 M & 75.30 & 92.20 \\
+SRM & 25.56 M & 77.51 & 93.56 \\
+GSoP & 28.20 M & 77.72 & 93.76 \\ 
+ECA & 25.56 M & 77.48 & 93.68 \\
+AutoLA & 29.39 M & \textbf{78.18} & 93.95 \\ 
+GCT & 25.56 M & 77.55 & 93.71 \\
+MCA & 25.56 M & 76.61 & 93.21 \\
+EAN & 25.56 M & 77.87 & \textbf{93.87} \\ 
+SASE & 28.10 M & 77.79 & 93.75 \\
\hline
ResNet-101 & 44.55 M & 76.7 & 93.12 \\
+SRM & 44.55 M & 78.76 & 93.92 \\
+GSoP & 48.90 M & 78.82 & 94.37 \\
+ECA & 44.55 M & 78.65 & 94.34 \\
+AutoLA & 51.81 M & 79.05 & 94.39 \\
+GCT & 44.55 M & 78.85 & 94.41 \\
+MCA & 44.68 M & 78.21 & 93.88 \\
+EAN & 44.55 M & 78.92 & 94.03 \\
+\textbf{SASE} & 49.36 M & \textbf{79.21} & \textbf{94.51} \\
\hline
\end{tabular}
\caption{The comparison results of image classification on ImageNet-1k validation set with other attention modules}
\label{table3}
\end{table}

\subsection{Object Detection on COCO}
For object detection, we selected Faster RCNN \cite{ren2015faster} and Mask RCNN \cite{he2017mask} as detectors with the modified ResNet-50 and ResNet-101 as backbones.
After the training on ImageNet was finished, we used the weights from the best-performing epoch as the pretrained backbone weights, and fine-tuned the whole network using COCO training set.
Each model was trained for 12 epochs.
For detectors with ResNet-50 as backbone, batch sizes were set to 8 with an initial learning rate of 0.01.
For detectors with ResNet-101 as backbone, batch sizes were set to 4 so that each model could fit into a single GPU.
The initial learning rate was set to 0.005 according to the linear scaling rule.
Learning rates were decayed by a factor of ten at the 8th and 11th epoch.
We applied SGD optimizer with a momentum of 0.9 and a weight decay of 0.0001.
Training techniques such as data pre-processing and FrozenBN, among others, were all implemented following mmdetection \cite{mmdetection} default settings.

Comparing to AutoLA \cite{ma2020auto}, SASE shows adequate amount of gain in overall AP performance across different backbones, indicating that searching in a more fine-grained manner indeed result in higher performance.
Comparing to other benchmark attention modules, our method also shows performance improvement on both Faster RCNN and Mask RCNN, demonstrating the ability of SASE to find better combinations of certain squeezing and excitation operations than existing algorithms.

\subsection{Instance Segmentation on COCO}
For instance segmentation, we evaluated SASE on Mask RCNN with the modified ResNet-50 and ResNet-101 as backbones.
All the training and validation settings followed the one from object detection.
SASE also outperforms all its counterparts on instance segmentation, therefore proving the effectiveness of our method across various tasks and networks.

\subsection{Ablation Study}
In order to find out how much performance improvement comes from the specific combination of operations, and how much is coming from the structure of SASE, we ran a random search on SASE and evaluated it using the same training and validation settings as introduced in above subsections.
The performance is compared with other attention blocks as well as the resulting architecture derived from careful searching.

For image classification, accuracy from random searching is worse than those of most attention methods on ResNet-50, but higher than those of most methods on ResNet-101.
Similar pattern is also found on object detection, where the relative ranking of random searching among other methods are higher on ResNet-101 and lower on ResNet-50.
This pattern reveals that the structure of SASE brings larger gain on deeper architectures than shallow ones.
We speculate that deeper networks generate richer feature representation and require more complicated way of processing them.
The structure of SASE makes use of both channel-wise and spatial-wise attention, which extract and process global information from different dimensions with different methods.
Such structure could handle more complex feature representation, and therefore provides more performance gain to deeper networks.

The resulting attention module obtained from architecture search outperforms the one from random search across all tasks and backbones by a notable margin.
Specifically, 0.55\% and 0.26\% gain on top-1 accuracy at ImageNet-1k validation set are observed, and \(0.6 \sim 0.9\% \) enhancement in mAP at COCO validation set are observed.
The ablation results demonstrate that SASE could further improve performance on multiple tasks when the architecture search is enabled.

\begin{table}[t]
\centering
\begin{tabular}{c|c|cc}
\hline SASE (random)+ & Params & Top-1 acc. & Top-5 acc. \\
\hline 
ResNet50 & 25.59 M & 77.22 & 93.48 \\
\hline
ResNet101 & 44.62 M & 78.95 & 94.46 \\
\hline
\end{tabular}
\caption{The ablation results of random searching on ImageNet-1k validation set}
\label{table5}
\end{table}

\begin{table}[t]
\centering
\begin{tabular}{c|c|cc}
\hline SASE (random)+ & Params & mAP & $AP_{0.5}$ \\
\hline
Faster RCNN res50 & 41.8 M & 38.6 & 60.3 \\
\hline
Faster RCNN res101 & 60.8 M & 41.1 & 62.3 \\
\hline
Mask RCNN res50 & 44.4 M & 39.1 & 60.5 \\
\hline
Mask RCNN res101 & 63.5 M & 41.4 & 62.6 \\
\hline
\end{tabular}
\caption{The ablation results of random searching on object detection on COCO 2017val}
\label{table5}
\end{table}


\section{Conclusion}
In summary, drawing inspiration from the common design pattern found in various attention blocks, we propose SASE, a novel search architecture tailored for squeeze-and-excitation style attention mechanisms. 
This architecture comprises separate function sets for squeeze and excitation operations across different dimensions.
Through applying differentiable search on the architecture, we obtained an attention module.
The searched module outperforms hand-crafted attention methods on a wide range of tasks and backbones, showing the effectiveness of our algorithm as well as its outstanding generalization ability.





\bibliography{aaai25}

\end{document}